\documentclass[runningheads]{llncs}
\usepackage{amsmath,amssymb} 

\usepackage{graphicx}

\usepackage{tikz}
\usepackage{comment}
\usepackage{color}

\usepackage[accsupp]{axessibility}  

\usepackage{microtype}
\usepackage{graphicx}
\usepackage{booktabs} 
\usepackage{multirow}

\usepackage{mathtools}
\usepackage{subcaption}
\usepackage{adjustbox}
\usepackage{gensymb}
\usepackage{hyperref}

\usepackage[capitalize,noabbrev]{cleveref}

\usepackage[textsize=tiny]{todonotes}
\usepackage{wrapfig}

\usepackage{tikz}
\def\checkmark{\tikz\fill[scale=0.4](0,.35) -- (.25,0) -- (1,.7) -- (.25,.15) -- cycle;}

\usepackage{bm}
\usepackage{bbm}
\newcommand{\vect}[1]{\boldsymbol{\mathbf{#1}}}
\DeclareMathOperator*{\argmax}{argmax}
\newcommand{\ubold}[1]{\fontseries{b}\selectfont#1}

\def\eg{\textit{e.g.}}
\def\ie{\textit{i.e.}}
\newcommand{\x}{\mathbf{x}}
\newcommand{\y}{\mathbf{y}}
\newcommand{\z}{\mathbf{z}}
\newcommand{\param}[1]{\vect{\theta}_{#1}}

\begin{document}
\pagestyle{headings}
\mainmatter
\def\ECCVSubNumber{2943}  

\title{Trust, but Verify: Using Self-Supervised Probing to Improve Trustworthiness} 

\titlerunning{Trust, but Verify: Using Self-Supervised Probing to Improve Trustworthiness}
%

\author{Ailin Deng \and
Shen Li \and
Miao Xiong \and
Zhirui Chen \and
Bryan Hooi
}
%
\authorrunning{A Deng et al.}
%

\institute{National University of Singapore \\
\email{\{ailin, shen.li, miao.xiong, zhiruichen\}@u.nus.edu }\\
\email{bhooi@comp.nus.edu.sg}}
\maketitle


\begin{abstract}
Trustworthy machine learning is of primary importance to the practical deployment of deep learning models. While state-of-the-art models achieve astonishingly good performance in terms of accuracy, recent literature reveals that their predictive confidence scores unfortunately cannot be trusted: e.g., they are often overconfident when wrong predictions are made, or so even for obvious outliers. In this paper, we introduce a new approach of \emph{self-supervised probing}, which enables us to check and mitigate the overconfidence issue for a trained model, thereby improving its trustworthiness. We provide a simple yet effective framework, which can be flexibly applied to  existing trustworthiness-related methods in a plug-and-play manner. Extensive experiments on three trustworthiness-related tasks (misclassification detection, calibration and out-of-distribution detection) across various benchmarks verify the effectiveness of our proposed probing framework.
\end{abstract}

\section{Introduction}
\label{sec:intro}

Deep neural networks have recently exhibited remarkable performance across a broad spectrum of applications, including image classification and object detection. However, the ever-growing range of applications of neural networks has also led to increasing concern about the reliability and trustworthiness of their decisions~\cite{toreini2020relationship,hendrycks2021unsolved}, especially in safety-critical domains such as autonomous driving and medical diagnosis. This concern has been exacerbated with observations about their overconfidence, where a classifier tends to give a wrong prediction with high confidence~\cite{nguyen2015deep,hendrycks17baseline}. Such disturbing phenomena are also observed on out-of-distribution data~\cite{hein2019relu}. The overconfidence issue thus poses great challenges to the application of models in the tasks of misclassification detection, calibration and out-of-distribution detection~\cite{hendrycks17baseline,guo2017calibration,hein2019relu}, which we collectively refer to as \emph{trustworthiness}.

Researchers have since endeavored to mitigate this overconfidence issue by deploying new model architectures under the Bayesian framework so as to yield well-grounded uncertainty estimates~\cite{Blundell2015WeightUI,kendall2017uncertainties,gal2016dropout}.  However, these proposed frameworks usually incur accuracy drops and heavier computational overheads. Deep ensemble models~\cite{lakshminarayanan2017simple,chen2021detecting} obtain uncertainty estimates from multiple classifiers, but also suffer from heavy computational cost. Some recent works~\cite{jiang2018trust,corbiere2019addressing} favor improving misclassification detection performance given a trained classifier. In particular, Trust Score~\cite{jiang2018trust} relies on the training data to estimate the test sample's misclassification probability, while True Class Probability~\cite{corbiere2019addressing} uses an auxiliary deep model to predict the true class probability in the original softmax distribution. Like these methods, our approach is plug-and-play and does not compromise the performance of the classifier, or require retraining it. Our method is complementary to existing trustworthiness methods, as we introduce the use of \emph{probing} as a new source of information that can be flexibly combined with trustworthiness methods.

Probing~\cite{alain2016understanding,hewitt2019designing} was proposed as a general tool for understanding deep models without influencing the original model. Specifically, probing is an analytic framework that uses the representations extracted from an original model to train another classifier (termed `probing classifier') for a certain designed probing task to predict some properties of interest. The performance (\eg, accuracy) of the learned probing classifier can be used to evaluate or understand the original model. For example, one probing framework proposed in~\cite{hewitt2019designing} evaluates the quality of sentence embeddings using the performance on the probing task of sentence-length or word-order prediction, while \cite{alain2016understanding} uses probing to understand the dynamics of intermediate layers in a deep network.

Though probing has been utilized in natural language processing for linguistics understanding, the potential of probing in mitigating the overconfidence issue in deep visual classifiers remains unexplored. Intuitively, our proposed framework uses probing to `assess' a classifier, so as to distinguish inputs where it can be trusted from those where it cannot, based on the results on the probing task. To achieve this goal, we need both 1) well-designed probing tasks, which should be different but highly related to the original classification task, and 2) a framework for how to use the probing results.

\begin{figure}[t]
    \centering
    \begin{subfigure}[b]{.22\textwidth}
    \centering
    \includegraphics[width=\linewidth]{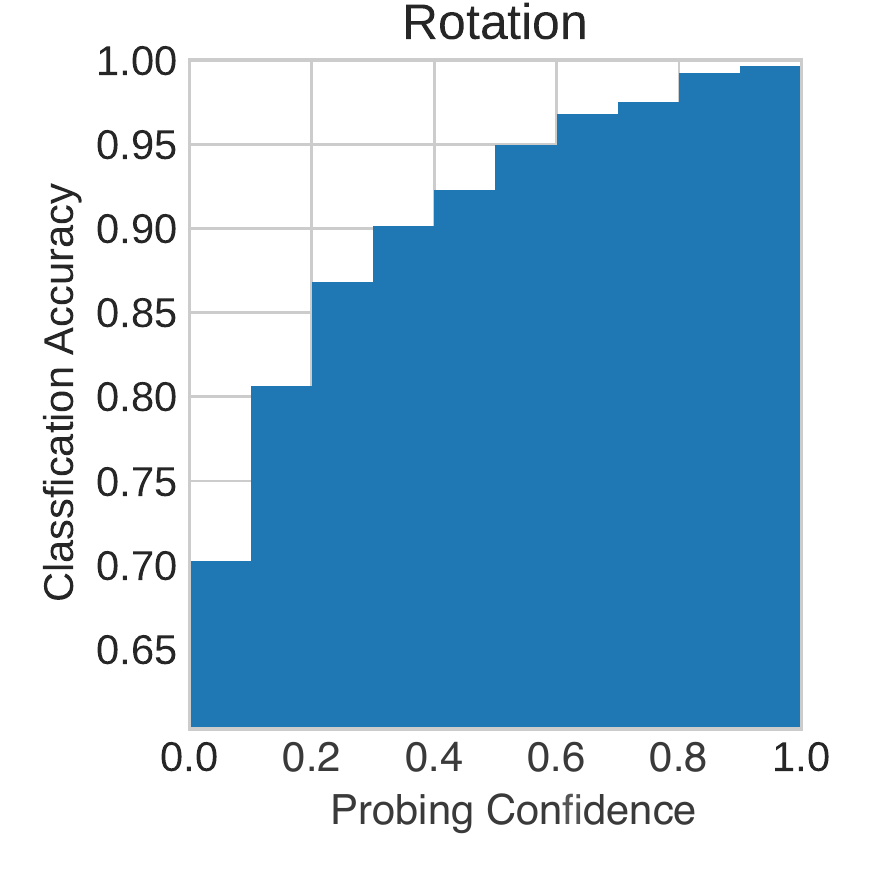}
    \label{fig:demo_probing_conf_correlation}
    \end{subfigure}%
    \begin{subfigure}[b]{.7\textwidth}
        \centering
        \includegraphics[width=\linewidth]{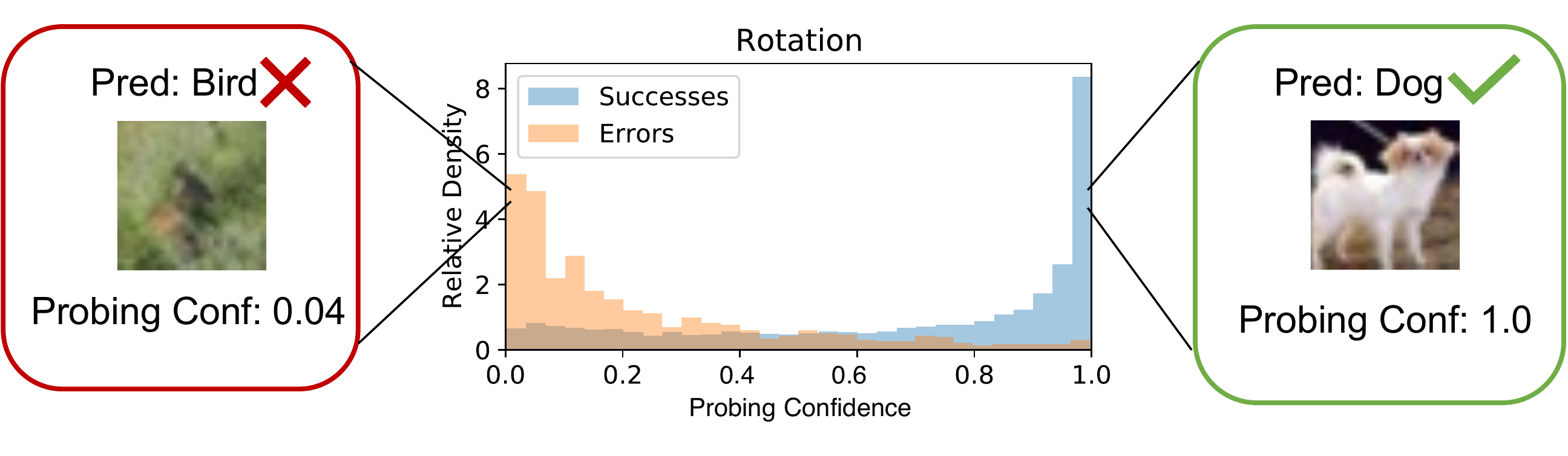}
        \label{fig:demo_probing_conf_dist}
    \end{subfigure}
    \caption{\emph{Left}: Positive correlation between probing confidence and classification accuracy. \emph{Right}: Images with lower probing confidence can be visually hard to detect and cause errors, while the images with higher probing confidences are sharp and clear, and lead to successful predictions. The visualized probing confidence is calculated from the self-supervised rotation task on CIFAR-10. }
    \label{fig:demo_probing_conf}
\end{figure}

So the first question is: \emph{what probing tasks are related to the original visual classification problem yet naturally available and informative}? We relate this question with the recent advancement of self-supervised learning. Existing literature suggests that a model that can tell the rotation, colorization or some other properties of objects is expected to have learned semantic information that is useful for downstream object classification~\cite{doersch2015unsupervised,noroozi2016unsupervised,gidaris2018unsupervised,zhang2016colorful}. Reversing this, we may expect that a pretrained supervised model with good classification performance can tell the properties of an object, \eg, rotation degrees. In addition, recently \cite{deng2021does} observes a strong correlation between rotation prediction accuracy and classification accuracy at the dataset level, under a multi-task learning scheme. This observation suggests that self-supervised tasks, \eg, rotation or translation prediction, can help in assessing a model's trustworthiness.

In our work, we first present a novel empirical finding that the `probing confidence', or the confidence of the probing classifier, highly correlates with the classification accuracy, as shown in Figure \ref{fig:demo_probing_conf} and Figure \ref{fig:probing_conf_acc}. Motivated by this finding, we propose our \emph{self-supervised probing} framework, which exploits the probing confidence for trustworthiness tasks, in a flexible and plug-and-play manner. Finally, we verify the effectiveness of our framework by conducting experiments on the three trustworthiness-related tasks.

Overall, the contributions and benefits of our approach are as follows\footnote{Our code is available at \url{ https://github.com/d-ailin/SSProbing}.}:
\begin{itemize}
    \item (Empirical Findings) We show that the probing confidence highly correlates with classification accuracy, showing the value of probing confidence as an auxiliary information source for trustworthiness tasks.
    \item (Generality) We provide a simple yet effective framework to incorporate the probing confidence into existing trustworthiness methods without changing the classifier architecture.
    \item (Effectiveness) We verify that our self-supervised probing framework achieves generally better performance in three trustworthiness related problems: misclassification detection, calibration and OOD detection.
\end{itemize}

\section{Related Work}
\label{sec:background}

\subsection{Trustworthiness in Deep Learning}

The overconfidence issue~\cite{hendrycks17baseline,hein2019relu} raises major concerns about deep models' trustworthiness, and has been studied in several related problems: calibration, misclassification detection, and out-of-distribution (OOD) detection~\cite{hendrycks17baseline,guo2017calibration,hein2019relu}.

Calibration algorithms aim to align a model's predictive confidence scores with their ground truth accuracy. Among these methods, a prominent approach is to calibrate the confidence without changing the original prediction, such as Temperature Scaling~\cite{guo2017calibration} and Histogram Binning~\cite{brier1950verification}.

For misclassification and OOD detection, a common approach is to incorporate uncertainty estimation to get a well-grounded confidence score. For example, \cite{malinin2018predictive,charpentier2020posterior} attempt to capture the uncertainty of every sample using a Dirichlet distribution. 
Ensemble-based methods such as Monte-Carlo Dropout~\cite{gal2016dropout} and Deep Ensembles~\cite{lakshminarayanan2017simple} calculate uncertainty from multiple trials either with the Bayesian formalism or otherwise. However, these uncertainty estimation algorithms have a common drawback that they involve modifying the classification architecture, thus often incurring accuracy drops. Besides, ensembling multiple overconfident classifiers can still produce overconfident predictions.

The practical demand for uncertainty estimation on pretrained models has led to a line of research developing \emph{post-hoc} methods. Trust Score \cite{jiang2018trust} utilizes neighborhood information as a metric of trustworthiness, assuming that samples in a neighborhood are most likely to have the same class label. True Class Probability \cite{corbiere2019addressing} aims to train a regressor to capture the softmax output score associated with the true class.

Compared to these works, we introduce \emph{probing confidence} as a valuable additional source of information for trustworthiness tasks. Rather than replacing existing trustworthiness methods, our approach is complementary to them, flexibly incorporating them into our self-supervised probing framework.

\subsection{Self-Supervised Learning}

Self-supervised learning leverages supervisory signals from the data to capture the underlying structure of unlabeled data. Among them, a prominent paradigm~\cite{doersch2015unsupervised,noroozi2016unsupervised,zhang2016colorful} is to define a prediction problem for a certain property of interest (known as pretext tasks) and train a model to predict the property with the associated supervisory signals for representation learning. For example, some works train models to predict any given image's rotation degree~\cite{gidaris2018unsupervised}, or the relative position of image patches~\cite{noroozi2016unsupervised}, or use multi-task learning combining supervised training with pretext tasks~\cite{hendrycks2019using}. The core intuition behind these methods is that the proposed pretext tasks are highly related to the semantic properties in images. As such, well-trained models on these tasks are expected to have captured the semantic properties in images. Motivated by this intuition but from an opposite perspective, we expect that the supervised models that perform well in object classification, should have grasped the ability to predict relevant geometric properties of the data, such as rotation angle and translation offset.

\subsection{Probing in Neural Networks}
Early probing papers~\cite{kohn2015s,sohn2015learning} trained `probing classifiers' on static word embeddings to predict various semantic properties. This analytic framework was then extended to higher-level embeddings, such as sentence embedding~\cite{adi2016fine} and contextual embedding~\cite{tenney2019you}, by developing new probing tasks such as predicting the properties of the sentence structure (\eg, sentence length) or other semantic properties. Apart from natural language processing, probing has also been used in computer vision to investigate the dynamics of each intermediate layer in the neural network~\cite{alain2016understanding}. However, most probing frameworks are proposed as an explanatory tool for analyzing certain characteristics of learned representations or models. Instead, our framework uses the probing framework to mitigate the overconfidence issue, by using the probing results to distinguish samples on which the model is trustworthy, from samples on which it is not.

\section{Methodology}
\label{sec:meth}

\subsection{Problem Formulation}
Let us consider a dataset $\mathcal{D}$ which consists of $N$ $\text{i.i.d}$ training samples, \ie, $\mathcal{D} = {(\vect{x}^{(i)}, y^{(i)})}_{i=1}^{N}$ where $\vect{x}^{(i)} \in \mathcal{R}^d $ is the $i$-th input sample and $y^{(i)} \in \mathcal{Y} = \{1, \ldots, K\}$ is the corresponding true class.

A classification neural network consists of two parts: the backbone parameterized by $\vect{\theta}_b$ and the linear classification layer parameterized by $\vect{\theta}_c$. Given an input $\x$, the neural network obtains a latent feature vector $\z = f_{\param{b}}(\x)$ followed by the softmax probability output and the predictive label:
\begin{align}
    \hat{P}(Y \mid \x, \param{b}, \param{c}) = \mathsf{softmax}(f_{\param{c}}(\z)) \\
    \hat{y} = \argmax_{k \in \mathcal{Y} } \hat{P}(Y =k \mid \x, \param{b}, \param{c}).
\end{align}

The obtained maximum softmax probability (MSP) $\hat{p} \coloneqq \hat{P}(Y = \hat{y} \mid \x, \param{b}, \param{c})$ is broadly applied in the three trustworthiness tasks: misclassfication detection, out-of-distribution detection and calibration~\cite{hendrycks17baseline,guo2017calibration}.

\subsubsection{Misclassification Detection}

is also known as error or failure prediction \cite{hendrycks17baseline,corbiere2019addressing}, and aims to predict whether a trained classifier makes an erroneous prediction for a test example. 
In general, it requires a confidence estimate for any given sample's prediction, where a lower confidence indicates that the prediction is more likely to be wrong.


For a standard network, the baseline method is to use the maximum softmax output as the confidence estimate for misclassification detection \cite{guo2017calibration,hendrycks17baseline}:

\begin{align}
 \hat{P}(\hat{y} \neq y) \coloneqq 1 - \hat{p}.
\end{align}
\subsubsection{Out-Of-Distribution Detection}
aims to detect whether a test sample is from a distribution that is different or semantically shifted from the training data distribution \cite{yang2021generalized}. 
\cite{hendrycks17baseline} proposed to use the maximum softmax scores for OOD detection. By considering the out-of-distribution data to come from a class that is not in $\mathcal{Y}$ (e.g. class $K+1$), we can write this as:
\begin{align}
 \hat{P}(y \in \mathcal{Y}) \coloneqq \hat{p},
\end{align}
where $y$ is the true label for sample $\mathbf{x}$. The minimum value of this score is $1/K$, so an ideal classifier when given out-of-distribution data is expected to assign a flat softmax output probability of $1/K$ for each class \cite{malinin2018predictive}.

\subsubsection{Calibration}
aims to align the predicted confidence $\hat{P}$ with the ground truth prediction accuracy. For example, with a well-calibrated model, samples predicted with confidence of $0.9$ should be correctly predicted $90\%$ of the time. Formally, we define perfect calibration as 
$$
    \mathbb{P}(\hat{Y} = Y \mid \hat{P} = p) = p, \ \forall\ p \in [0,1],
$$
where $\hat{P}$ is estimated as the maximum softmax probability in the standard supervised multi-class classification setup. However, these scores have commonly been observed to be miscalibrated, leading to a line of research into calibration techniques for neural networks, such as \cite{guo2017calibration,Mukhoti2020CalibratingDN}. 

On the whole, the baseline confidence scores (MSP) $\hat{p}$ have been observed to be often overconfident, even on misclassified as well as out-of-distribution samples \cite{guo2017calibration,hendrycks17baseline,hein2019relu}. This degrades the performance of the baseline approach on all three tasks: misclassification detection, OOD detection and calibration. Our work aims to show that self-supervised probing provides a valuable source of auxiliary information, which helps to mitigate the overconfidence issue and improve performance on these three tasks in a post-hoc setting.

\begin{figure}[t]
    \centering
    \includegraphics[width=\linewidth]{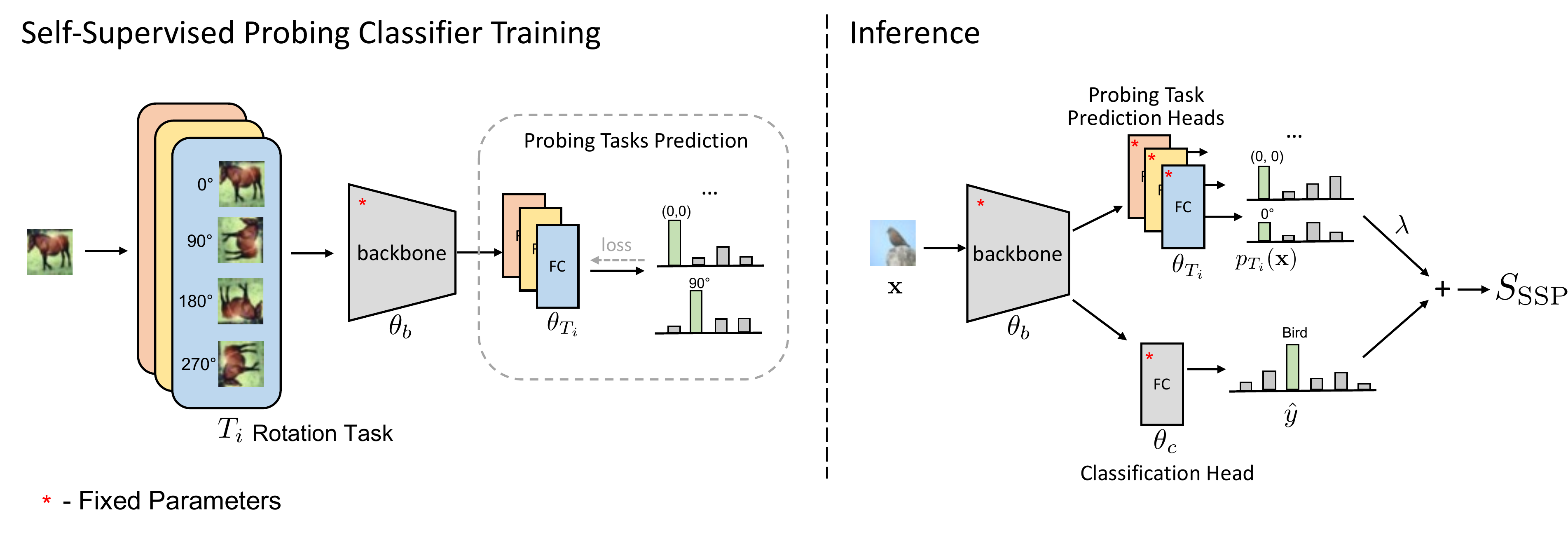}
    \caption{Our self-supervised probing framework, which first trains a probing classifier (left); then at test time, combines the probing confidence with the confidence obtained from the classifier.}
    \label{fig:framework}
\end{figure}

\subsection{Self-Supervised Probing Framework}

\subsubsection{Overview.} 
Our self-supervised probing framework computes the \emph{probing confidence}, and uses it as an auxiliary source of information for the three trustworthiness tasks, given a trained classifier. Our framework involves two steps:
\begin{enumerate}
\item Training the self-supervised probing classifier to obtain the probing confidence for each sample;
\item Incorporating probing confidence into the three trustworthiness tasks. Specifically, for misclassification and OOD detection, we incorporate probing confidence by combining it with the original confidence scores. For calibration, we propose a simple and novel scheme which uses the probing confidence as prior information for input-dependent temperature scaling.
\end{enumerate} 
This framework is illustrated in Figure~\ref{fig:framework}.

\subsubsection{Self-Supervised Probing Tasks.}

Recall that our goal is to use probing tasks to assess the trustworthiness of the classifier. This requires probing tasks that are semantically relevant to the downstream classification task (but without using the actual class labels). The observations made in  \cite{gidaris2018unsupervised,zhang2016colorful,noroozi2016unsupervised}

suggest that simple tasks which apply a discrete set of transformations (e.g. a set of rotations or translations), and then require the model to predict which transformation was applied, should be suitable as probing tasks.

Formally, we denote the set of probing tasks as $\mathcal{T} = \{ T_{1}, T_{2}, \ldots, T_{M} \}$, where each task $T_i$ consists of $k_i$  transformations $T_{i} = \{ t_i^{(0)}, t_i^{(1)}, \ldots, t_i^{(k_i-1)} \}$, where $t_i^{(0)}$ is the identity transformation. 
For example, one can create a rotation probing task defined by four rotation transformations associated with rotation degrees of $\{ 0\degree, 90\degree, 180\degree, 270\degree \}$, respectively. 

\subsubsection{Training Probing Classifier.} As our goal is to provide auxiliary uncertainty support for a given model, we avoid modifying or fine-tuning the original model and fix the model's backbone throughout training. Thus, for a given probing task $T_i \in \mathcal{T}$, we fix the supervised model's backbone $f_{\param{b}}$ and train the \emph{probing classifier} as a fully-connected (FC) layer with parameters $\param{T_i}$. Optimization proceeds by minimizing the cross entropy loss $\mathcal{L}_{\text{CE}}$ over $\param{T_i}$ only:
\begin{align}
    \hat{P}(Y_{T_i} &\mid t(\x), \param{b}, \param{T_i}) := \mathsf{softmax}( f_{\param{T_i}}(f_{\param{b}}( t (\x)) ) \\
    \min_{\param{T_i}} \mathcal{L}_{T_i} &:= \mathbb{E}_{(\x, y) \sim \mathcal{D}} \sum_{t \in T_i} \mathcal{L}_{\text{CE}}(\y_{T_i}, \hat{P}(Y_{T_i} \mid t(\x), \param{b}, \param{T_i})),
\end{align}
where $\y_{T_i}$ is the one hot label for the probing task and $\mathcal{L}_{\text{CE}}$ denotes the cross entropy loss.

As the backbone is fixed for all probing tasks and there are no other shared parameters among probing tasks, the training for all probing tasks are performed in parallel. After training, we obtain $M$ probing classifiers for the probing tasks $\mathcal{T}$ ($|\mathcal{T}| = M$).

\subsubsection{Computing Probing Confidence.} 
During inference, for each test image $\x$ and probing task $T_i$, we will now compute the \emph{probing confidence} to help assess the model's trustworthiness on $\x$. Intuitively, if the model is trustworthy on $\x$, the probing classifier should correctly recognize that $\x$ corresponds to an identity transformation (since it is the original untransformed test image). Thus, we probe the model by first passing the test image through the backbone, followed by applying the probing classifier corresponding to task $T_i$. Then, we compute the probing confidence $p_{T_i}(\x) \in \mathbb{R}$ as the probing classifier's predictive confidence for  the identity transformation label (i.e. for label $0$) in the softmax probability distribution:
\begin{align}
    p_{T_i}(\x) := \hat{P}(\y_{T_{i}}^{(0)} \mid \x, \param{b}, \param{T_{i}})
\end{align}

\subsubsection{Empirical Evidence.}

\begin{figure*}[t]
    \centering
    \includegraphics[width=.8\linewidth]{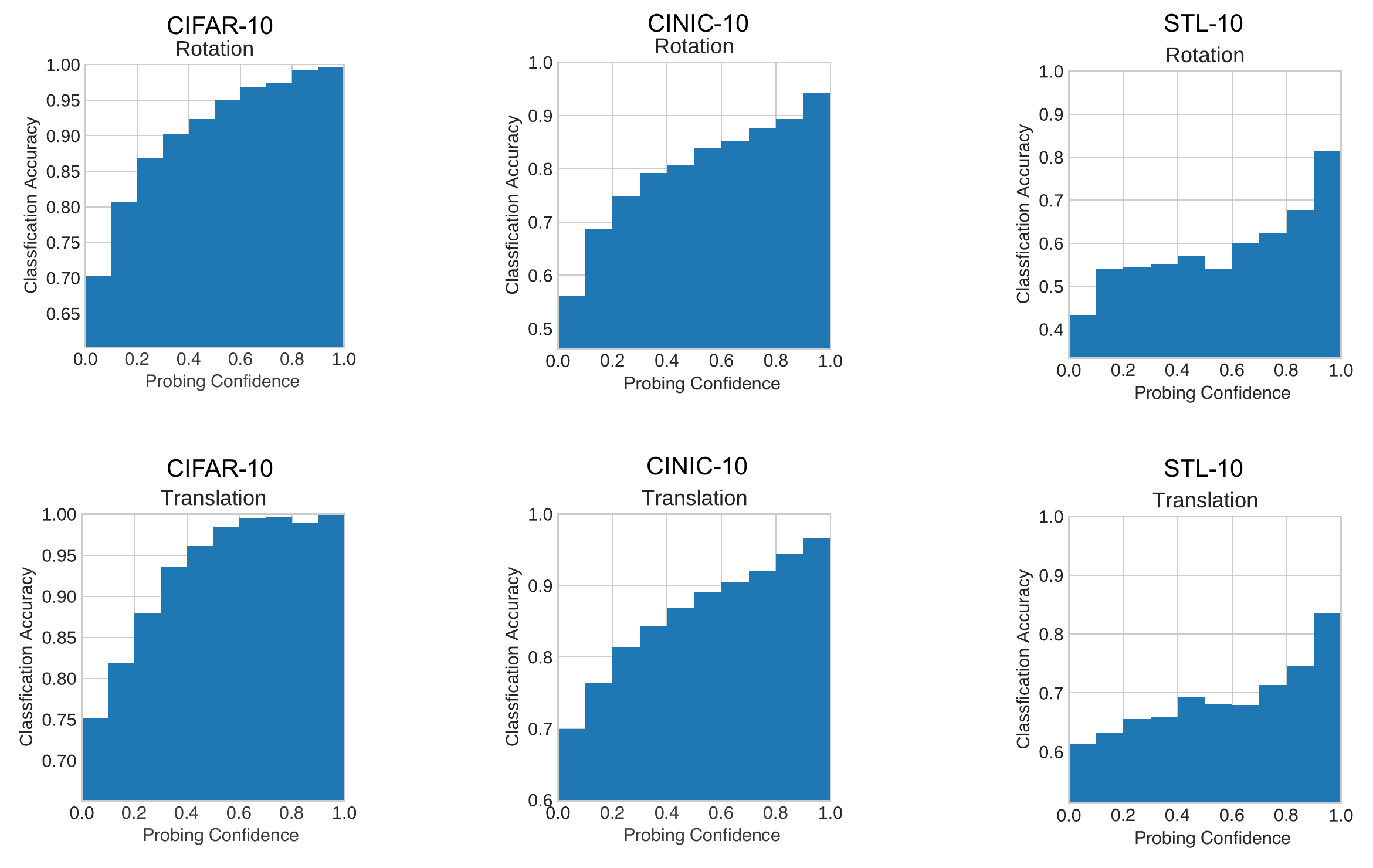}
    \caption{Clear positive correlation between classification accuracy and probing confidence under the rotation and translation probing tasks on CIFAR-10, CINIC-10 and STL-10. }
    \label{fig:probing_conf_acc}
\end{figure*}

From Figure \ref{fig:probing_conf_acc}, we observe that for both rotation and translation probing tasks, and on three datasets, the probing confidence has a clear positive correlation with the classification accuracy. This empirical evidence indicates that the samples with higher probing confidence tend to be predicted correctly in the classification task. This validates our use of probing confidence for assessing predictive confidence given a sample.

\subsubsection{Incorporating Probing Confidence.}
\label{sec:probing_conf}

For the misclassification detection task, we compute our \emph{self-supervised probing} score by combining the probing confidence from the different probing tasks with any existing misclassification score $S(\x)$, which can be the classifier's maximum softmax probability, entropy, or any other existing indicator scores~\cite{jiang2018trust,corbiere2019addressing}:

\begin{align}
    S_{\text{SSP}}(\x) := S(\x) + \sum_{i=1}^{M} \lambda_{i} p_{T_i}(\x),
    \label{eq:ssl_weighted_score}
\end{align}

where $\mathbf{\lambda}_i$’s are hyperparameters and are determined corresponding to the best AUPR-ERR performance on the validation set. The proposed $S_{\text{SSP}}(\x)$ is the combined result from the original indicator and the probing confidence scores.

Similarly, we do the same for OOD detection, where $S(\x)$ can be any existing OOD score, e.g. maximum softmax probability or entropy.

\subsubsection{Input-Dependent Temperature Scaling.}
\label{sec:input_temp_scaling}
For the calibration task, we design our input-dependent temperature scaling scheme to calibrate the original predictive confidence as an extension of temperature scaling~\cite{guo2017calibration}. Classical temperature scaling uses a single scalar temperature parameter $a_0$ to rescale the softmax distribution. Using our probing confidence $p_{T_i}(\x)$ for each sample $\x$ as prior information, we propose to obtain a scalar temperature $\tau(\x)$ as a learned function of the probing confidence: 
\begin{align}
    \tau(\x) &:= a_0 + \sum_{i=1}^{M} a_{i} p_{T_i}(\x) \\
    \tilde{P}(Y \mid \x) &:= \mathsf{softmax} \left(\frac{f_{\param{c}} \left(f_{\param{b}}(\x)\right)} {\tau(\x)}\right)  
    \label{eq:ssl_cal}
\end{align}
Here, $\tilde{P}(Y \mid \x)$ contains our output calibrated probabilities. $a_0$ and $a_i$ are learnable parameters; they are optimized via negative likelihood loss on the validation set, similarly to in classical temperature scaling~\cite{guo2017calibration}.
For each sample $\x$, we obtain $\tau(\x)$ as its input-dependent temperature. With $\tau(\x) = 1$, we recover the original predicted probabilities $\hat{p}$ for the sample. As all logit outputs of a sample are divided by the same scalar, the predictive label is unchanged. In this way, we calibrate the softmax distribution based on the probing confidence, without compromising the model's accuracy.

\section{Experiments}
\label{sec:exp}

In this section, we conduct experiments on the three trustworthiness-related tasks: misclassification, OOD detection and calibration. The main results including ablation study and case study focus on the misclassification detection task, while the experiments on calibration and OOD performance aim to verify the general effectiveness of probing confidence for trustworthiness-related tasks. 

\subsection{Experimental Setup}
\subsubsection{Datasets.} We conduct experiments on the benchmark image datasets: CIFAR-10~\cite{torralba200880}, CINIC-10~\cite{darlow2018cinic} and STL-10~\cite{coates2011analysis}. We use the default validation split from CINIC-10 and split $20\%$ data from the labeled training data as validation set for CIFAR-10 and STL-10. All the models and baselines share the same setting. Further details about these datasets, architectures, training and evaluation metrics can be found in the supplementary material.  

\subsubsection{Network Architectures.}
Our classification network architectures adopt the popular and effective models including VGG16~\cite{simonyan2014very} and ResNet-18~\cite{he2016deep}. For fairness, all methods share the same classification network. We train each probing task with a FC layer.
The hyperparameters $\lambda_i$’s are selected according to the best AUPR-ERR performance on the corresponding validation set.

\subsubsection{Evaluation Metrics.}
The evaluation metrics for misclassification, OOD detection and calibration follow the standard metrics used in the literature~\cite{hendrycks17baseline,corbiere2019addressing,guo2017calibration}. We relegate the details to the supplementary material.

\subsection{Results on Misclassification Detection}

\begin{table*}[t]
  \caption{Comparison of misclassification detection methods. All methods share the same trained classification network. All values are percentages.  \text{+SSP} indicates incorporating our self-supervised probing. \textbf{Bold} numbers are the superior results.}
  \label{tab:misclassification}
  \centering
  \begin{adjustbox}{max width=.9\textwidth}
  \begin{tabular}{cl*{4}c}
    \toprule
     &  & FPR@95\% $\downarrow$ & AUPR-ERR $\uparrow$ & AUPR-SUCC $\uparrow$ & AUROC $\uparrow$ \\
     \cmidrule(r){3-6}
    Dataset & Model & \multicolumn{4}{c}{Base/+SSP} \\
    \midrule
    \multirow{4}{*}{\shortstack[c]{\ubold{CIFAR-10} \\ VGG16}} 
    & MSP & 50.50/\textbf{48.87} & 46.21/\textbf{46.98} & 99.13/\textbf{99.16} & 91.41/\textbf{91.53} \\
    & MCDropout & 50.25/\textbf{49.37} & 46.64/\textbf{47.23} & 99.15/\textbf{99.17} & 91.46/\textbf{91.58} \\
    & TCP & 45.74/\textbf{45.61} & 47.70/\textbf{47.93} & 99.16/\textbf{99.19} & 91.85/\textbf{91.88} \\
    & TrustScore & 47.87/\textbf{45.61} & 46.50/\textbf{47.65} & 98.99/\textbf{99.16} & 90.47/\textbf{91.68} \\
    \midrule
    \multirow{4}{*}{\shortstack[c]{\ubold{CIFAR-10} \\ ResNet-18}} 
    & MSP & 47.01/\textbf{45.30} & 44.39/\textbf{45.48} & 99.02/\textbf{99.32} & 90.63/\textbf{92.10} \\
    & MCDropout & 43.47/\textbf{38.18} & 42.04/\textbf{52.60} & \textbf{99.53}/99.51 & 93.09/\textbf{94.05} \\
    & TCP & \textbf{40.88}/\textbf{40.88} & \textbf{50.37} /50.36 & \textbf{99.48}/\textbf{99.48} & \textbf{93.74}/93.73 \\
    & TrustScore  & 31.62/\textbf{30.77} & 59.57/\textbf{60.12} & 99.46/\textbf{99.47} & 94.29/\textbf{94.38} \\

    \midrule
    \multirow{4}{*}{\shortstack[c]{\ubold{CINIC-10} \\ VGG16}} 
    & MSP & 67.23/\textbf{66.49} & 53.21/\textbf{54.40} & \textbf{96.67}/96.11 & \textbf{86.33}/85.53 \\
    & MCDropout & 64.74/\textbf{64.62} & 54.02/\textbf{54.76} & 94.41/\textbf{96.19} & 84.85/\textbf{86.45} \\
    & TCP & 67.80/\textbf{65.95} & 53.19/\textbf{54.27} & 96.57/\textbf{96.58} & 86.51/\textbf{86.62} \\
    & TrustScore  & 68.36/\textbf{65.65} & 51.83/\textbf{53.66} & 96.19/\textbf{96.46} & 85.25/\textbf{86.01} \\
    
    \midrule
    \multirow{4}{*}{\shortstack[c]{\ubold{CINIC-10} \\ ResNet-18}} 
    & MSP & 62.57/\textbf{62.48} & 53.18/\textbf{53.29} & \textbf{97.73}/97.57 & \textbf{88.39}/88.04 \\
    & MCDropout & 59.32/\textbf{58.21} & 52.55/\textbf{57.20} & \textbf{98.23}/\textbf{98.23} & 89.50/\textbf{90.16} \\
    & TCP & 59.66/\textbf{58.95} & 55.08/\textbf{55.27} & 97.87/\textbf{97.89} & 89.07/\textbf{89.17} \\
    & TrustScore  & 62.26/\textbf{60.08} & 53.06/\textbf{54.53} & 97.64/\textbf{97.73} & 88.07/\textbf{88.50} \\
    
    \midrule
    \multirow{4}{*}{\shortstack[c]{\ubold{STL-10} \\ ResNet-18}} 
    & MSP & 77.12/\textbf{76.67} & 58.81/\textbf{59.19} & 89.59/\textbf{89.81} & 78.99/\textbf{79.35} \\
    & MCDropout & 74.09/\textbf{73.86} & 60.49/\textbf{61.01} & \textbf{92.20}/91.33 & 81.55/\textbf{81.59} \\
    & TCP & 79.19/\textbf{79.07} & 54.72/\textbf{54.94} & 85.18/\textbf{85.59} & 74.79/\textbf{75.17} \\
    & TrustScore  & 72.48/\textbf{71.99} & 61.36/\textbf{62.10} & 90.60/\textbf{90.81} & 80.57/\textbf{80.95} \\
    \bottomrule
  \end{tabular}
  \end{adjustbox}
\end{table*}

\subsubsection{Performance.} To demonstrate the effectiveness of our framework, we implemented the baseline methods including Maximum Softmax Probabilty~\cite{hendrycks17baseline}, Monte-Carlo Dropout (MCDropout)~\cite{gal2016dropout}, Trust Score~\cite{jiang2018trust} and True Class Probability (TCP)~\cite{corbiere2019addressing}. Our implementation is based on the publicly released code (implementation details can be found in supplementary materials). To show the effectiveness of probing confidence for these existing trustworthiness scores, we compare the performance of models with and without our self-supervised probing approach (refer to Eq. \eqref{eq:ssl_weighted_score}, where we use the existing baseline methods as $S(\x)$).

The results are summarized in Table \ref{tab:misclassification}. From the table, we observe that our method outperforms baseline scores in most cases. This confirms that probing confidence is a helpful indicator for failure or success prediction, and improves the existing state-of-the-art methods in a simple but effective way. 


\subsubsection{Q1: When does self-supervised probing adjust the original decision to be more (or less) confident? }

As our goal is to provide auxiliary evidence support for predictive confidence based on self-supervised probing tasks, we investigate what kinds of images are made more or less confident by the addition of the probing tasks.

\begin{figure}[h]
    \centering
    \includegraphics[width=\linewidth]{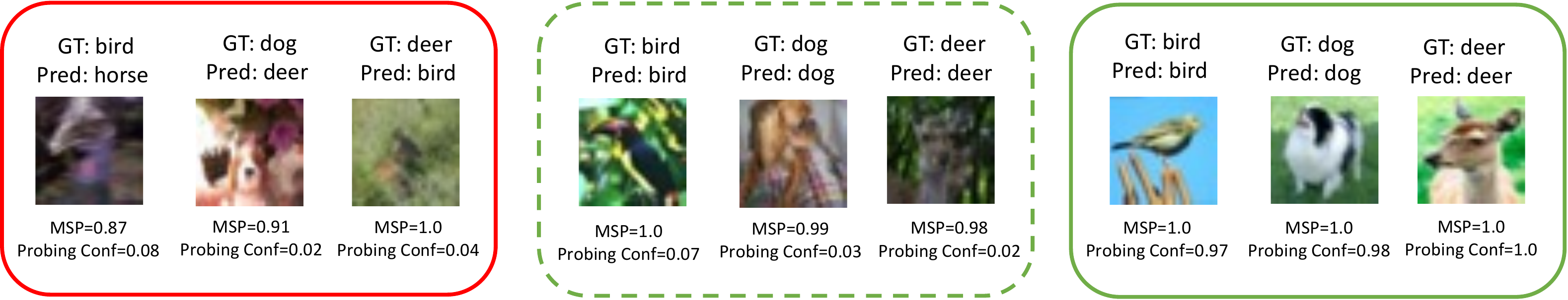}
    \caption{Analysis of the examples in CIFAR-10 for the rotation probing task. The red box contains the samples that are wrongly predicted on both probing and classification tasks. The green dashed box contains the samples predicted wrongly on the probing task but predicted correctly for classification. The green solid box contains the samples predicted correctly on both probing tasks and classification. The objects that are visually harder to detect tend to fail in the probing task. }
    \label{fig:case_samples}
\end{figure}

Figure \ref{fig:case_samples} illustrates three cases, demonstrating what kind of images tend to fail or succeed in the probing tasks. From the figure, we observe that samples with objects that are intrinsically hard to detect (e.g., hidden, blurred or blending in with their surroundings)  tend to fail in the probing task, whereas samples with clear and sharp objects exhibit better performance in the probing task. The former type of samples are likely to be less trustworthy, which validates the intuition for our approach.

\subsubsection{Q2: How do different combinations of probing tasks affect performance?}

To further investigate the effect of the probing tasks, we design different combinations of probing tasks and observe how these combinations affect the performance in misclassification tasks.

We first demonstrate the performance with or without rotation and translation probing tasks to see how each task can affect the performance. The result is based on the ResNet-18 model trained on CIFAR-10 and reported in Table \ref{tab:ablation_comb}. The result shows that the rotation probing task contributes more than the translation task to the overall improvement. The combination of rotation and translation probing tasks outperforms each one individually, implying that  multiple probing tasks better identifies the misclassified samples, by combining different perspectives.
\begin{table}
    \centering
    \caption{The performance of different combinations of probing tasks for CIFAR-10. Combining both probing tasks outperforms the individual task setting, suggesting that multiple tasks effectively assesses trustworthiness from multiple perspectives. }

        \centering
        \adjustbox{max width=.75\linewidth}{
        \begin{tabular}{c c c c c c}
            \toprule
            & & \multicolumn{4}{c}{CIFAR-10} \\
             \cmidrule(r){3-6}
             Rotation & Translation & FPR@95\% $\downarrow$ & AUPR-ERR $\uparrow$ & AUPR-SUCC $\uparrow$ & AUROC $\uparrow$  \\
             \midrule
             \checkmark & \checkmark & \textbf{45.30} & \textbf{45.48} & \textbf{99.32 } & \textbf{92.10} \\
             \checkmark &  & 45.58 & 45.36 & 99.28 & 91.78 \\
              & \checkmark & 46.15 & 44.61 & 99.24 & 91.42 \\
             \bottomrule
        \end{tabular}}
    \label{tab:ablation_comb}
\end{table}

To further investigate the influence of the number of transformations for each probing task, we conduct experiments by varying the numbers of transformations in the probing task. 

\begin{figure}[]
    \centering
    \begin{subfigure}[b]{.35\textwidth}
    \centering
    \includegraphics[width=\linewidth]{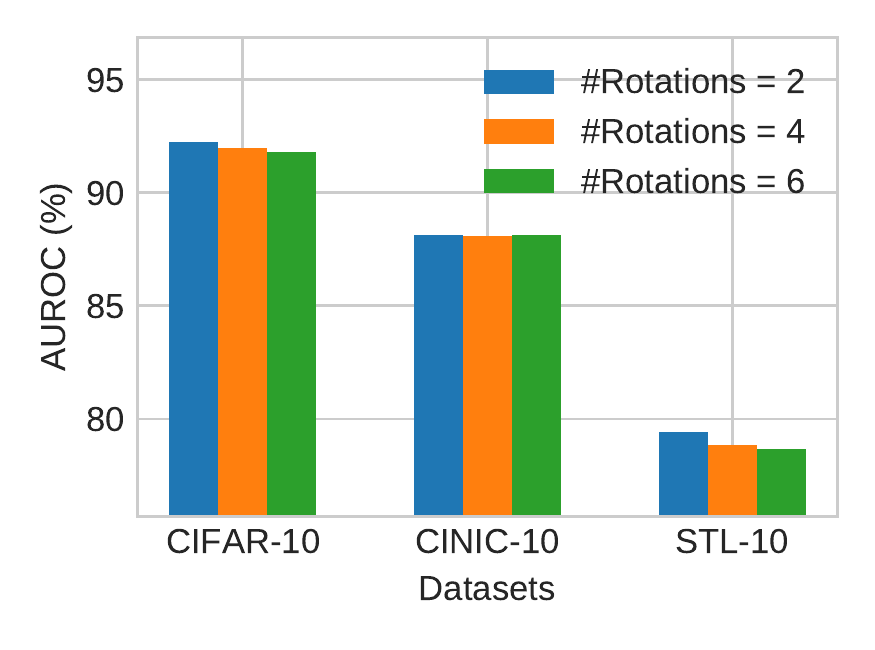}
    \label{fig:ablation_rot_num}
    \end{subfigure}%
    \begin{subfigure}[b]{.35\textwidth}
        \centering
        \includegraphics[width=\linewidth]{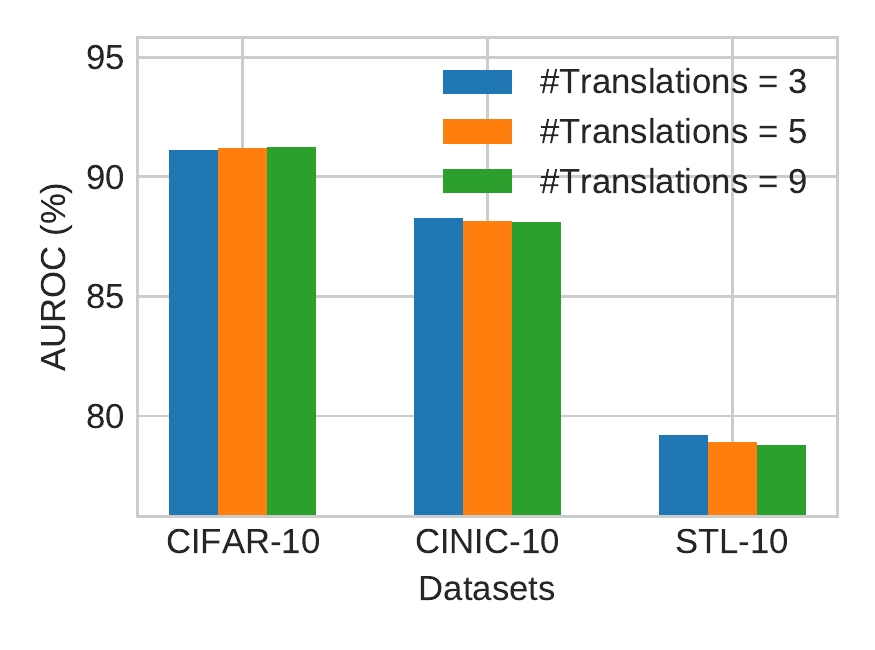}
        \label{fig:ablation_trans_num}
    \end{subfigure}
    \caption{The performance (AUROC) when using different numbers of transformations (\#Rotations / \#Translations) in the rotation and translation probing tasks. The probing tasks with more transformations decrease the  performance in most case, especially for the small dataset (STL-10). }
    \label{fig:ablation_complexity}
\end{figure}

The details of the experimental setting are provided in the supplementary material and the result is shown in Figure \ref{fig:ablation_complexity}. 

We observe that the larger dataset (CINIC-10) shows stable performance under the probing tasks with varying number of transformations, but the smaller dataset (STL-10) shows a drop in performance when using the probing tasks with more transformations. As the number of transformations in a probing task can be regarded as the complexity of the probing task, this suggests that on smaller datasets, the probing tasks should be designed with fewer transformations to allow the probing classifier to effectively learn the probing task.

\subsubsection{Q3: Feasibility of other self-supervised tasks than rotation and translation prediction.}
\label{sec:sub:task_set}
Other than rotation and translation, there are self-supervised tasks such as  jigsaw puzzles~\cite{noroozi2016unsupervised}, \ie, predicting the jigsaw puzzle permutations. Since our self-supervised probing framework can extract the probing confidence for any proposed self-supervised probing tasks flexibly, we also experiment with jigsaw puzzle prediction as a probing task. However, we found that the training accuracy for jigsaw puzzle prediction is low, resulting in less informative probing confidence scores. This is probably because shuffling patches of an image breaks down the image semantics, making it challenging for the supervised backbone to yield meaningful representations for the self-supervised probing task. 

In general, probing tasks should be simple yet closely related to visual semantic properties, so that the probing confidence correlates with classification accuracy. The rotation and translation tasks assess a model's ability to identify the correct orientation and position profile of the object of interest, which are closely related to the classification task; but more complex tasks (e.g., jigsaw) can lead to greater divergence between probing and classification. We leave the question of using other potential probing tasks in the further study.

\subsection{Results on Out-of-Distribution Detection}
Besides misclassification detection, we also conduct experiments on out-of-distribution detection with $S_{\text{SSP}}$. All hyperparameters $\lambda_i$ share the same setting as in misclassification detection. Since our goal is to verify that our self-supervised probing approach can be combined with common existing methods to enhance their performance, we build upon the most commonly used methods for OOD detection: Maximum Softmax Probability (MSP) and the entropy of the softmax probability distribution (refer to Eq. \eqref{eq:ssl_weighted_score}).

The results are reported with the AUROC metric in Table \ref{tab:ood}, indicating that our self-supervised probing consistently improves the OOD detection performance on both MSP and entropy methods.

\begin{table}[t]
\caption{AUROC (\%) of OOD detection trained on in-distribution data (a) CIFAR-10 and (b) CINIC-10. The baseline methods are Maximum Softmax Probability and Entropy. \text{+SSP} indicates incorporating our self-supervised probing.}
\label{tab:ood}
\begin{subtable}[h]{\textwidth}
  \label{ood-results:cifar10}
  \centering
  \begin{adjustbox}{max width=.78\textwidth}
  \begin{tabular}{ll*{5}c}
    \toprule

    & & \multicolumn{5}{c}{CIFAR-10 $\to$} \\
    \cmidrule(r){3-7}
    Backbone & Method & SVHN & LSUN & ImageNet & LSUN(FIX) & ImageNet(FIX) \\
    \midrule
    \multirow{4}{*}{VGG16} 
     & MSP & 91.05 & 88.98 & 88.08 & 85.64 & 86.01 \\
     & MSP+SSP & \textbf{92.57} & \textbf{90.39} & \textbf{90.08} & \textbf{86.93} & \textbf{87.41} \\
     \cmidrule(r){2-7}
     & Entropy & 91.79 & 89.55 & 88.60 & 86.04 & 86.46\\
     & Entropy+SSP & \textbf{92.57} & \textbf{90.70} & \textbf{90.09} & \textbf{86.76} & \textbf{87.34} \\
     \midrule
     \multirow{4}{*}{ResNet-18} 
     & MSP & 88.80 & 91.17 & 88.86 & 85.26 & 85.75 \\
     & MSP+SSP & \textbf{91.71} & \textbf{92.62} & \textbf{91.27} & \textbf{89.71} & \textbf{89.65} \\          \cmidrule(r){2-7}
     & Entropy & 89.27 & 91.94 & 89.44 & 85.54 & 86.06 \\
     & Entropy+SSP & \textbf{92.13} & \textbf{93.55} & \textbf{92.02} & \textbf{90.04} & \textbf{90.03} \\
  \end{tabular}
  \end{adjustbox}
\end{subtable}
\hfill
\begin{subtable}[h]{\linewidth}
 
  \label{ood-results:cinci10}
  \centering
  \begin{adjustbox}{max width=.75\textwidth}
  \begin{tabular}{ll*{5}c}
    \toprule

    & & \multicolumn{5}{c}{CINIC-10 $\to$} \\
    \cmidrule(r){3-7}
    Backbone & Method & SVHN & LSUN & ImageNet & LSUN(FIX) & ImageNet(FIX) \\
    \midrule
    \multirow{4}{*}{VGG16} 
     & MSP & 81.48 & 81.17 & 80.36 & 76.45 & 76.76 \\
     & MSP+SSP & \textbf{85.25} & \textbf{84.73} & \textbf{83.53} & \textbf{80.92} & \textbf{80.36} \\
     \cmidrule(r){2-7}
     & Entropy & 83.24 & 82.96 & 81.92 & 77.83 & 77.93 \\
     & Entropy+SSP & \textbf{85.07} & \textbf{84.57} & \textbf{83.32} & \textbf{79.90} & \textbf{79.68} \\
     \midrule
     \multirow{4}{*}{ResNet-18} 
     & MSP & 88.81 & 86.11 & 83.03 & 83.25 & 81.59 \\
     & MSP+SSP & \textbf{90.33} & \textbf{87.87} & \textbf{86.20} & \textbf{83.69} & \textbf{83.43} \\
      \cmidrule(r){2-7}
     & Entropy & 87.65 & 83.34 & 80.96 & 83.59 & 81.34 \\
     & Entropy+SSP & \textbf{89.75} & \textbf{85.46} & \textbf{83.26} & \textbf{85.09} & \textbf{83.31} \\
    \bottomrule
  \end{tabular}
  \end{adjustbox}
\end{subtable}

\end{table}

\subsection{Results on Calibration}
In this section, we verify our proposed input-dependent temperature scaling as described in Section \ref{sec:input_temp_scaling}. Specifically, we compare the common calibration baselines, including Temperature Scaling~\cite{guo2017calibration} and Histogram Binning~\cite{zadrozny2001obtaining}. Temperature Scaling is the key baseline for verifying the effectiveness of our use of probing confidence as prior information to obtain a temperature for each sample.

The result is shown in Table \ref{tab:calibrate}. We observe that our proposed calibration method generally outperforms the baseline methods under different evaluation metrics.

\begin{table*}[t]
  \caption{The reported performance in calibration. Our approach (Scaling+SSP) compare with uncalibrated softmax probability (MSP)~\cite{hendrycks17baseline}, Histogram Binning binning (Hist. Binning)~\cite{zadrozny2001obtaining} and Temperature Scaling (Temp. Scaling)~\cite{guo2017calibration}.}
  \label{tab:calibrate}
  \centering
  \begin{adjustbox}{max width=.78\textwidth}
  \begin{tabular}{cl*{4}c}
    \toprule
     &  & ECE (\%) $\downarrow$ & MCE (\%) $\downarrow$ & NLL $\downarrow$ & Brier Score ($\times 10^{-3}$) $\downarrow$ \\
    \midrule
    \multirow{4}{*}{\shortstack[c]{\ubold{CIFAR-10} \\ VGG16}} 
    & MSP (uncalibrated) & 5.0 & 31.43 & 0.39 & 13.17 \\
    & Hist. Binning  & 1.65 & 20.68 & 0.35 & 12.88   \\
    & Temp. Scaling  & 1.03 & \textbf{7.53} & \textbf{0.26} & \textbf{12.03}  \\
    & Scaling+SSP & \textbf{0.93} & 9.15 & \textbf{0.26} & \textbf{12.03} \\
    \midrule
    \multirow{5}{*}{\shortstack[c]{\ubold{CIFAR-10} \\ ResNet-18}} 
    & MSP (uncalibrated) & 4.31 & 28.16 & 0.28 & 11.55 \\
    & Hist. Binning & 1.17 & 27.95 & 0.31 & 10.73  \\
    & Temp. Scaling  & 1.40 & 18.91 & 0.23 & 10.72 \\
    & Scaling+SSP & \textbf{0.75} & \textbf{7.92} & \textbf{0.22} & \textbf{10.48} \\

    \midrule
    \multirow{5}{*}{\shortstack[c]{\ubold{CINIC-10} \\ VGG16}} 
    & MSP (uncalibrated) & 9.68 & 24.29 & 0.71 & 27.82 \\
    & Hist. Binning & 2.95 & 28.40 & 0.67 & 26.44  \\
    & Temp. Scaling & 0.62 & \textbf{2.46} & \textbf{0.55} & 25.34 \\
    & Scaling+SSP & \textbf{0.53} & 3.42 & \textbf{0.55} & \textbf{25.28} \\
    \midrule
    \multirow{5}{*}{\shortstack[c]{\ubold{CINIC-10} \\ ResNet-18}} 
    & MSP (uncalibrated)  & 7.94 & 23.08 & 0.55 & 23.43 \\
    & Hist. Binning & 2.26 & 21.09 & 0.56 & 22.20   \\
    & Temp. Scaling & 1.41 & 13.30 & 0.45 & 21.56 \\
    & Scaling+SSP & \textbf{0.77} & \textbf{10.22} & \textbf{0.44} & \textbf{21.51} \\
    \midrule
    \multirow{5}{*}{\shortstack[c]{\ubold{STL-10} \\ ResNet-18}} 
    & MSP (uncalibrated) & 16.22 & 26.76 & 1.18 & 46.73 \\
    & Hist. Binning & 7.80 & 17.73 & 1.92 & 46.30   \\
    & Temp. Scaling & 1.56 & 9.08 & \textbf{0.89} & 42.22 \\
    & Scaling+SSP & \textbf{1.17} & \textbf{7.61} & \textbf{0.89} & \textbf{42.15} \\
    \bottomrule
  \end{tabular}
  \end{adjustbox}
\end{table*}

\section{Conclusions}
\label{sec:concl}

In this paper, we proposed a novel self-supervised probing framework for enhancing existing methods' performance on trustworthiness related problems. We first showed that the `probing confidence' from the probing classifier highly correlates with classification accuracy. Motivated by this, our framework enables incorporating probing confidence into three trustworthiness related tasks: misclassification, OOD detection and calibration. We experimentally verify the benefits of our framework on these tasks. 
Our work suggests that self-supervised probing serves as a valuable auxiliary information source for trustworthiness tasks across a wide range of settings, and can lead to the design of further new methods incorporating self-supervised probing (and more generally, probing) into these and other tasks, such as continual learning and open-world settings.

\paragraph{Acknowledgements.} This work was supported in part by NUS ODPRT Grant R252-000-A81-133.

\bibliographystyle{splncs04}


\end{document}


\pagestyle{headings}
\mainmatter
\def\ECCVSubNumber{2943}  

\title{Supplementary Material of Trust, but Verify: Using Self-Supervised Probing to Improve Trustworthiness} 

\titlerunning{Trust, but Verify: Using Self-Supervised Probing to Improve Trustworthiness}
%
\author{Ailin Deng \and
Shen Li \and
Miao Xiong \and
Zhirui Chen \and
Bryan Hooi
}
%
\authorrunning{A Deng et al.}
%

\institute{National University of Singapore \\
\email{\{ailin, shen.li, miao.xiong, zhiruichen\}@u.nus.edu }\\
\email{bhooi@comp.nus.edu.sg}}
\maketitle










\appendix


\section{Experimental Setup}
We implement the based models and self-supervised probing framework in Pytorch (GPU) 1.10.1+cu113 and train them on a server with Intel(R) Xeon(R) Gold 6226R CPU @ 2.90GHz and a GeForce RTX 3090 graphics card. The implementation is based on the publicly released codes\footnote{\url{https://github.com/google/TrustScore}\newline \url{https://github.com/valeoai/ConfidNet}\newline \url{https://github.com/markus93/NN\_calibration}}.

\section{Dataset}
We conduct experiments on the benchmark image datasets: CIFAR-10~\cite{torralba200880}, CINIC-10~\cite{darlow2018cinic} and STL-10~\cite{coates2011analysis}. We use the default validation split from CINIC-10 and split $20\%$ data from the labeled training data as the validation sets for CIFAR-10 and STL-10, respectively. All the models and baselines share the same dataset setting for fair comparison. 
\begin{itemize}
    \item CIFAR-10~\cite{torralba200880}: CIFAR-10 consists of 50000 training images and 10000 testing images at a resolution of $32\times32$. The dataset covers ten object classes, with each class having an equal number of images. We split 10000 images out of the training images as the validation set.
    \item CINIC-10~\cite{darlow2018cinic}: CINIC-10 dataset is designed to be a middle option relative to CIFAR-10 and ImageNet~\cite{deng2009imagenet}: it contains images at a resolution of $32\times32$ as CIFAR10 but at a large scale total of 270000 images, which is closer to that of ImageNet. The dataset has default data splits: equal 90000 images for training, validation and test set. 
    \item STL-10: STL-10 is an image dataset derived from ImageNet with a resolution of $96\times96$. It contains 100000 unlabeled images and 13000 labeled images from 10 object classes. Among the labeled images, 5000 images are partitioned for training while the remaining 8000 images are for testing. We split 1000 images for each class from the training images as validation data.
\end{itemize}

\section{Classification Network Architecture}
For classification, we adopted the architectures VGG16~\cite{simonyan2014very} and ResNet-18~\cite{he2016deep}. The specific architectures we used are publicly released\footnote{\scriptsize \url{https://github.com/valeoai/ConfidNet/blob/master/confidnet/models/vgg16.py}\newline\url{https://github.com/weiaicunzai/pytorch-cifar100/blob/master/models/resnet.py}}. To enable comparing with the MCdropout baseline, we add dropout after each block in the ResNet models. The VGG16 models are trained using the Adam optimizer with learning rate $1 \times 10^{-4}$ and $(\beta_1, \beta_2) = (0.9, 0.99)$. As the adopted public VGG16 models can not support the images with $96 \times 96$ resolution directly, we only train VGG16 for CIFAR-10 and CINIC-10. The ResNet models are trained using the SGD optimizer with cosine annealing scheduler beginning with learning rate of $0.1$. The models for CIFAR-10 and CINIC-10 are trained for 300 epochs and the model for STL-10 are trained for 100 epochs as training with the dataset at a smaller scale can reach convergence faster. We choose the model at the final epoch as our base trained classification model. The test accuracies for the trained model are reported in the Table \ref{tab:acc}.


\begin{table}[]
    \centering
    \caption{Test accuracies (\%) for each trained classification model.}
    \label{tab:acc}
    \adjustbox{width=\linewidth}{
    \begin{tabular}{lccccc}
        \toprule
        & \shortstack[c]{\ubold{CIFAR-10} \\ VGG16} & \shortstack[c]{\ubold{CIFAR-10} \\ ResNet-18} & \shortstack[c]{\ubold{CINIC-10} \\ VGG16} & \shortstack[c]{\ubold{CINIC-10} \\ ResNet-18} & \shortstack[c]{\ubold{STL-10} \\ ResNet-18} \\
        \midrule
        Test Accuracy (\%) & 92.02 & 93.34 & 82.10 & 84.75 & 69.30 \\
        \bottomrule
    \end{tabular}}
    
\end{table}


\section{Evaluation}
\subsection{Misclassification Detection}

\subsubsection{FPR at 95\% TPR} measures the False Positive Rate (FPR) when the True Positive Rate (FPR) is equal to 95\%. True Positive Rate is computed by $\texttt{TPR} = \texttt{TP}/(\texttt{TP} + \texttt{FN})$, where $\texttt{TP}$ and $\texttt{FN}$ denote the occurrences of true positives and false negatives, respectively. The False Positive Rate can be computed by $\texttt{FPR} = \texttt{FP}/(\texttt{FP} + \texttt{TN})$, where $\texttt{FP}$ and $\texttt{TN}$ denote the occurrences of false positives and true negatives, respectively. One can interpret the metric as the probability that a sample is misclassified when the True Positive Rate (TPR) is equal to 95\%.

\subsubsection{AUROC} measures the Area Under the Receiver Operating Characteristic curve (AUROC). It is a threshold-agnostic performance evaluation metric, as the curve shows the trade-off between TPR and FPR across different decision thresholds.  

\subsubsection{AUPR} measures the Area Under the Precision-Recall (PR) curve. The PR curve is a graph showing precision = TP/(TP + FP) versus recall = TP/(TP + FN) across different decision thresholds. Similar to AUROC, AUPR is also a threshold-agnostic performance evaluation. In our tests, AUPR-SUC indicates that correct predictions are used as the positive class, while AUPR-ERR indicates that errors are used as the positive class.

For we train our probing framework only on train set and select hyperparameter $\lambda$ on validation set, our baselines are also implemented on train data except that we train TCP for CIFAR10 with validation set as TCP~\cite{corbiere2019addressing} relies on a larger absolute number of errors while our obtained classifiers can achieve nearly 100\% accuracy on a training set and get less errors compared to using validation set.


\subsection{OOD}
For Out-of-Distribution Detection (OOD), we use SVHN~\cite{netzer2011reading}, ImageNet~\cite{deng2009imagenet} and LSUN~\cite{yu2015lsun} -related datasets as OOD datasets and use CIFRA-10 and CINIC-10 as the normal datasets. We use SVHN from the Pytorch library; and all the other datasets are publicly released\footnote{\url{https://github.com/alinlab/CSI}}. As our goal is to verify that the self-supervised probing can benefit the existing commonly used OOD methods, we compare with the Maximum Softmax Probability (MSP) and the entropy. We adopt AUROC as our general OOD performance evaluation metric.

\subsection{Calibration}
\subsubsection{ECE} measures Expected Calibration Error. We partition predictions into $M$ equally spaced bins and compute the accuracy for each bin. ECE is the average of the bins' accuracy/confidence difference. We use $M = 15$ in our experiments. 
\subsubsection{MCE} measures Maximum Calibration Error. Unlike ECE, MCE is the largest calibration error across all bins. It measures the worse-case deviation between confidence and accuracy.
\subsubsection{NLL} is Negative Log Likelihood and a standard measure of a probabilistic model's quality~\cite{hastie2009elements}. It is also referred to as the cross entropy loss~\cite{lecun2015deep}. 
\subsubsection{Brier Score} is equivalent to the mean squared error when applied to predicted probabilities for unidimensional predictions. For multi-class classification, it can be defined as:
\begin{equation}
B S=\frac{1}{N} \sum_{t=1}^{N} \sum_{i=1}^{R}\left(f_{t i}-o_{t i}\right)^{2},
\end{equation}
where $R$ is the number of possible classes and $N$ denotes the total number of instances across all classes. $f_{t i}$ is the predictive probability of class $i$ and $o_{t i}$ is 1 if the example $t$ has a ground-truth class label $i$. 

\section{Self-Supervised Probing}

In our experiments, we design rotation and translation as the probing tasks. Specifically, we design rotation prediction tasks with 4 degrees $\{ 0\degree, 90\degree, 180\degree, 270\degree \}$ and translation tasks with 5 different translated pixels $\{ (0, 0), (-8, 0), (8, 0), (0, -8), (0, 8) \}$ for both CIFAR-10 and CINIC-10 across different backbone models. As for the smaller dataset STL-10, we design easier probing tasks: rotation prediction tasks with 2 degrees $\{ 0\degree, 90\degree \}$ and translation tasks with 3 different translated pixels $\{ (0, 0), (0, -32), (0, 32)\}$. To avoid the trivial solution for learning translation tasks, the translated images' paddings are interpolated under reflection mode. The probing classifiers are trained for $10$ epochs using the SGD optimizer and the training learning rate is scheduled with cosine annealing schedulers.

\section{More Experimental Results}
This section provides more experimental results as follows.
\subsection{The empirical evidence for probing confidence}
\begin{figure}
    \centering
    \includegraphics[width=\linewidth]{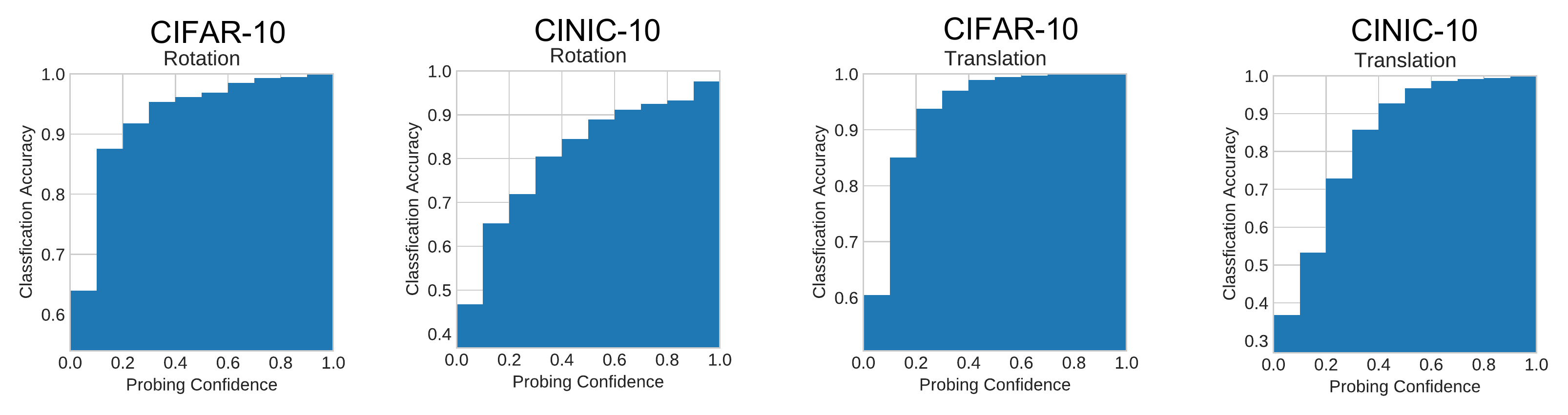}
    \caption{Clear positive correlation between classification accuracy and probing confidence under the rotation and translation probing tasks with VGG16 models trained on CIFAR-10 and CINIC-10, respectively.}
    \label{fig:app_correlation}
\end{figure}

Figure~\ref{fig:app_correlation} shows the correlation between probing confidence and accuracy with VGG16 backbones trained on the CIFAR-10 and CINIC-10 datasets. 

\begin{figure}[t]
    \centering
    \includegraphics[width=\linewidth]{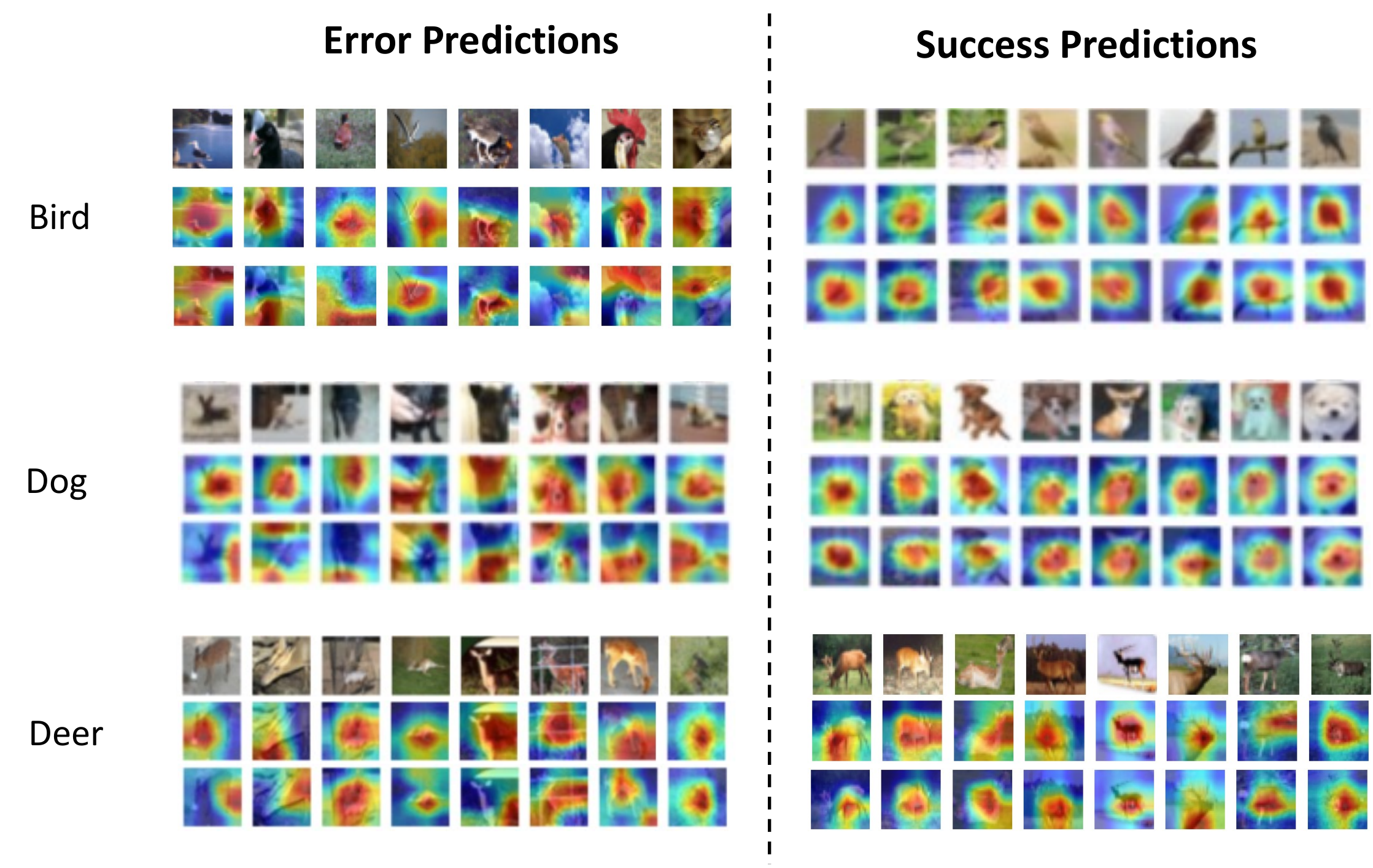}
    \caption{Error and success predictions with activation maps in object classification and probing tasks. For each sample, the first row shows the original images, the second row shows the activation maps for object classification and the third row demonstrates the activation maps for the probing task. The success predictions tend to have consistent regions in the activation maps, which cover the objects of interest in the images.}
    \label{fig:app_cases}
\end{figure}
\subsection{Q1: When does self-supervised probing adjust the original decision to be more (or less) confident?}
This section provides the qualitative analysis of our proposed framework. As observed in the main paper, the images with clear and sharp objects tend to succeed in both object classification and probing tasks, while the images with objects intrinsically hard to detect tend to fail in the probing tasks even for object classification. Going a step further, it is natural to ask whether the classifier has been able to correctly localize the objects in the images such that it can perform well in both object classification and probing classification. To answer it, we use a gradient-based visualization algorithm, Grad-Cam~\cite{selvaraju2017grad}\footnote{\url{https://github.com/jacobgil/pytorch-grad-cam}}, to show the activation maps for both object classification and probing task (rotation) classification. As demonstrated in Figure~\ref{fig:app_cases}, we observe that, for successful predictions, the resultant activation maps obtained from two tasks are consistent with the highlighted regions enveloping the objects of interest. In contrast, the misclassification samples tend to exhibit the activation maps of different highlighted regions for object classification and probing tasks.

\subsection{Q2: How do different combinations of probing tasks affect performance?}
We have conducted ablation study on the combinations of probing tasks. Specifically, we have designed varying numbers of transformations in a probing task. For example, for rotation task, we have designed 2, 4, 6 transformations: $\{0\degree, 90\degree\}$,  $\{0\degree, 90\degree, 180\degree, 270\degree\}$ and $\{0\degree, 60\degree, 120\degree, 180\degree, 240\degree, 300\degree\}$. For translation tasks, we have designed 3, 5, 9 transformations: $\{ (0, 0), (0, -8), (0, 8) \}$, $\{ (0, 0), (-8, 0), (8, 0),$ $(0, -8), (0, 8) \}$ and $\{ (0, 0), (-8, 0), (8, 0), (0, -8), (0, 8), (8, 8), (-8, -8), (8, -8), (-8, 8) \}$. The results are reported based on the ResNet-18 models trained on different datasets under misclassification detection settings. We report the results with AUROC (\%) in the Table \ref{tab:app_ablation}. 

\begin{table}[t]
    \caption{Results (AUROC \%) for different transformations in a probing task.}
    \label{tab:app_ablation}
    \centering
    \begin{tabular}{l|ccc}
        \toprule
         & CIFAR-10 & CINIC-10 & STL-10  \\
         \midrule
        \#rotations = 2 & 92.25 & 88.12 & 79.43 \\
        \#rotations = 4 & 91.97 & 88.10 & 78.87 \\
        \#rotations = 6 & 91.79 & 88.14 & 78.69 \\
        \midrule
        \#translations = 3 & 91.11 & 88.28 & 79.22 \\
        \#translations = 5 & 91.22 & 88.15 & 78.90 \\
        \#translations = 9 & 91.26 & 88.11 & 78.80 \\
        \bottomrule
    \end{tabular}
    
\end{table}

\subsection{Comparison with the baseline of using randomly initialized probing head.}
\begin{figure}[]
    \centering
    \includegraphics[width=.3\linewidth]{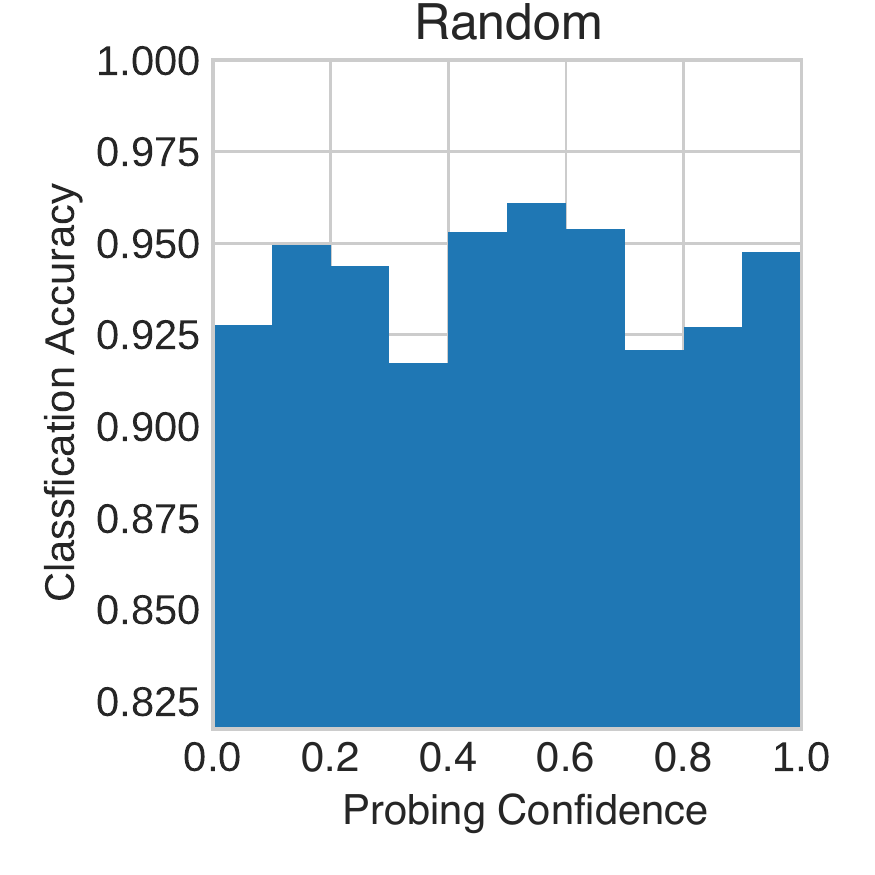}
    \caption{The correlation between classification accuracy and probing confidence with randomly initialized probing head.}
    \label{fig:app_random}
\end{figure}

Figure \ref{fig:app_random} displays the correlation between classification accuracy and probing confidence with randomly initialized head without any training. We obtained the probing confidence that is almost uniformly distributed and not correlated to the classification accuracy any more, which further verifies the nontrivial effectiveness of the SSL learned probing heads. We use normal distribution $\mathcal{N}\left(0, 1\right)$ for the random head initialization.

\bibliographystyle{splncs04}
\bibliography{BIB/trust,BIB/ssl,BIB/dataset,BIB/probe}
